\documentclass[10pt,twocolumn,letterpaper]{article}

\usepackage[pagenumbers]{cvpr} 

\usepackage{graphicx}
\usepackage{amsmath}
\usepackage{amssymb}
\usepackage{booktabs}
\usepackage{multirow}
\usepackage{algorithm}
\usepackage{algpseudocode}
\usepackage{pifont}

\usepackage[pagebackref,breaklinks,colorlinks]{hyperref}

\usepackage[capitalize]{cleveref}
\crefname{section}{Sec.}{Secs.}
\Crefname{section}{Section}{Sections}
\Crefname{table}{Table}{Tables}
\crefname{table}{Tab.}{Tabs.}

\def\cvprPaperID{} 
\def\confName{CVPR}
\def\confYear{2023}

\begin{document}

\title{Iterative Scene Graph Generation with Generative Transformers}

\author{Sanjoy Kundu\\
Department of Computer Science\\
Oklahoma State University\\
{\tt\small sanjoy.kundu@okstate.edu}
\and
Sathyanarayanan N. Aakur\\
Department of Computer Science\\
Oklahoma State University\\
{\tt\small saakurn@okstate.edu}
}
\maketitle

\begin{figure*}[th]
    \centering
    \includegraphics[width=0.99\textwidth]{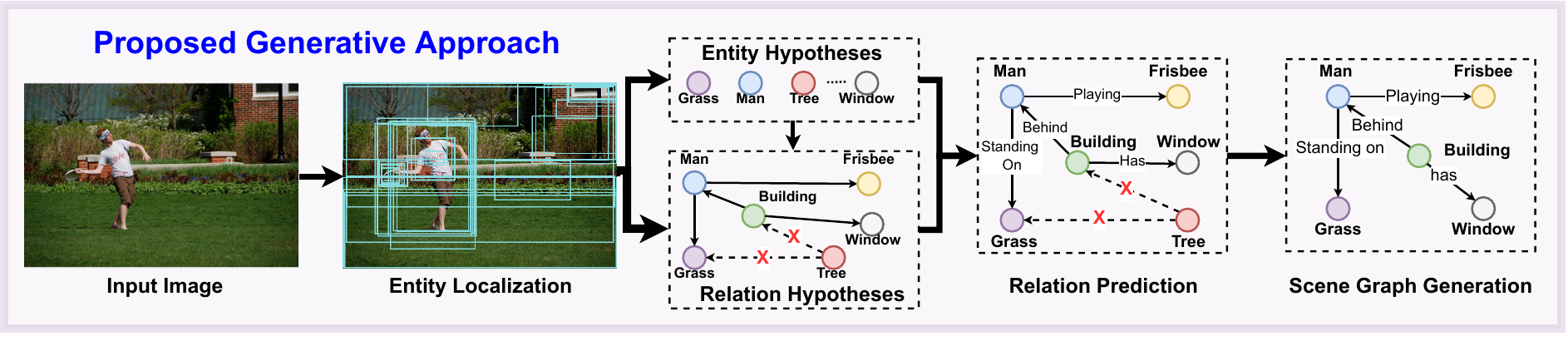}
    \caption{Our goal is to move towards a \textit{generative} model for scene graph generation using a two-stage approach where we first sample the underlying semantic structure between entities before predicate classification. This is different from the conventional approach of modeling pairwise relationships among all detected entities and helps constrain the reasoning to the underlying semantic structure.}
    \label{fig:intuition}
\end{figure*}

\begin{abstract}
Scene graphs provide a rich, structured representation of a scene by encoding the entities (objects) and their spatial relationships in a graphical format. This representation has proven useful in several tasks, such as question answering, captioning, and even object detection, to name a few. Current approaches take a generation-by-classification approach where the scene graph is generated through labeling of all possible edges between objects in a scene, which adds computational overhead to the approach. This work introduces a generative transformer-based approach to generating scene graphs beyond link prediction. Using two transformer-based components, we first sample a possible scene graph structure from detected objects and their visual features. We then perform predicate classification on the sampled edges to generate the final scene graph. This approach allows us to efficiently generate scene graphs from images with minimal inference overhead. Extensive experiments on the Visual Genome dataset demonstrate the efficiency of the proposed approach. Without bells and whistles, we obtain, on average, $20.7\%$ mean recall (mR@100) across different settings for scene graph generation (SGG), outperforming state-of-the-art SGG approaches while offering competitive performance to \textit{unbiased} SGG approaches. 
\end{abstract}

\section{Introduction}
\label{sec:intro}
Graph-based visual representations are becoming increasingly popular due to their ability to encode visual, semantic, and even temporal relationships in a compact representation that has several downstream tasks such as object tracking~\cite{bal2022bayesian}, scene understanding~\cite{johnson2015image} and event complex visual commonsense reasoning~\cite{aakur2019going,aakur2022knowledge,liang2022visual}. Graphs can help navigate clutter and express complex semantic structures from visual inputs to mitigate the impact of noise, clutter, and (appearance/scene) variability, which is essential in scene understanding. Scene graphs, defined as directed graphs that model the visual-semantic relationships among entities (objects) in a given scene, have proven to be very useful in downstream tasks such as visual question-answering~\cite{hildebrandt2020scene,teney2017graph}, captioning~\cite{johnson2015image}, and even embodied tasks such as navigation~\cite{ravichandran2022hierarchical}, to name a few. 

There has been a growing body of work~\cite{xu2017scene,zellers2018neural,Tang_2019_CVPR,Chen_2019_CVPR,yang2021probabilistic,guo2021general,shit2022relationformer} that has focused on the problem of scene graph generation (SGG), that aims to generate scene graph from a given input observation. However, such approaches have tackled the problem by beginning with a fully connected graph, where all entities interact with each other before pruning it down to a more compact graph by predicting edge relationships, or the lack of one, between each pair of localized entities. This approach, while effective, has several limitations. First, by modeling the interactions between entities with a dense topology, the underlying semantic structure is ignored during relational reasoning, which can lead to poor predicate (relationship) classification. Second, by constructing pairwise relationships between all entities in a scene, there is tremendous overhead on the predicate classification modules since the number of pairwise comparisons can grow non-linearly with the increase in the number of detected concepts. Combined, these two issues aggravate the existing long-tail distribution problem in scene graph generation. Recent progress in \textit{unbiasing}~\cite{Tang_2019_CVPR,tang2020unbiased,suhail2021energy,Li_2022_CVPR} has attempted to address this issue by tackling the long-tail distribution problem. However, they depend on the quality of the underlying graph generation approaches, which suffer from the above limitations.

In this work, we aim to overcome these limitations using a two-stage, generative approach called IS-GGT, a transformer-based iterative scene graph generation approach. An overview of the approach is illustrated in Figure~\ref{fig:intuition}. Contrary to current approaches to SGG, we leverage advances in generative graph models~\cite{liao2019efficient,belli2019image} to first sample the underlying interaction graph between the detected entities before reasoning over this sampled semantic structure for scene graph generation. By decoupling the ideas of graph generation and relationship modeling, we can constrain the relationship classification process to consider only those edges (pairs of entities) that have a higher probability of interaction (both semantic and visual) and hence reduce the computational overhead during inference. Additionally, the first step of generative graph sampling (Section~\ref{sec:ggt}) allows us to navigate clutter by rejecting detected entities that do not add to the semantic structure of the scene by iteratively constructing the underlying entity interaction graph conditioned on the input image. A relation prediction model (Section~\ref{sec:rel_pred}) reasons over this constrained edge list to classify the relationships among interacting entities. Hence, the relational reasoning mechanism only considers the (predicted) global semantic structure of the scene and makes more coherent relationship predictions that help tackle the long-tail distribution problem without additional unbiasing steps and computational overhead. 



\textbf{Contributions.} The contributions of this paper are three-fold: (i) we are among the first to tackle the problem of scene graph generation using a \textit{graph generative} approach without constructing expensive, pairwise comparisons between all detected entities, (ii) we propose the idea of iterative interaction graph generation and global, contextualized relational reasoning using a two-stage transformer-based architecture for effective reasoning over cluttered, complex semantic structures, and (iii) through extensive evaluation on Visual Genome~\cite{krishna2017visual} we show that the proposed approach achieves state-of-the-art performance (without unbiasing) across all three scene graph generation tasks while considering only $20\%$ of all possible pairwise edges using an effective graph sampling approach. 
\section{Related Work}
\label{sec:relatedWork}
\textbf{Scene graph generation}, introduced by Johnson \textit{et al.}~\cite{johnson2015image}, aims to construct graph-based representations that capture the rich semantic structure of scenes by modeling objects, their interaction and the relationships between them. Most approaches to scene graph generation have followed a typical pipeline: object detection followed by \textit{pairwise} interaction modeling to generate plausible \textit{(Subject, Predicate, Object)} tuples, which represent the labeled edge list of the scene graph. Entity localization (i.e., concept grounding) has primarily been tackled through localization and labeling of images through advances in object detection~\cite{ren2015faster,carion2020end}. The relationship or predicate classification for obtaining the edge list tuples has focused mainly on capturing the global and local contexts using mechanisms such as recurrent neural networks and graph neural networks to result in seminal approaches to scene graph generation such as IMP~\cite{xu2017scene}, MOTIFS~\cite{zellers2018neural}, and R-CAGCN~\cite{yang2018graph}. Single-stage methods such as FC-SSG~\cite{Liu_2021_CVPR} and Relationformer~\cite{shit2022relationformer}, as well relational modeling approaches such as RelTR~\cite{cong2022reltr} have integrated context through transformer-based ~\cite{vaswani2017attention} architectures. However, these approaches fail to explicitly tackle the long-tail distributions prevalent in visual scene graphs as proposed by Tang \textit{et al.}~\cite{Tang_2019_CVPR} and Chen \textit{et al.}~\cite{Chen_2019_CVPR}, concurrently. \textbf{Unbiased scene graph generation} models explicitly tackle this problem by building upon SGG models such as VCTree and MOTIFs to provide better predicate classification. Several approaches have been successfully applied to tackle unbiased generation, such as using external knowledge (VCTree~\cite{Tang_2019_CVPR} and KERN~\cite{Chen_2019_CVPR}), counterfactual reasoning (TDE~\cite{tang2020unbiased}), energy-based loss functions (EBML~\cite{suhail2021energy}), modeling predicate probability distributions (PPDL~\cite{Li_2022_CVPR} and PCPL~\cite{10.1145/3394171.3413722}), graphical contrastive losses~\cite{zhang2019graphical}, cognitive trees (CogTree~\cite{yu2020cogtree}, bi-level sampling~\cite{Li_2021_CVPR}, and regularized unrolling (RU-Net\cite{Lin_2022_CVPR}), to name a few. However, these approaches still perform expensive pairwise comparisons to obtain the final scene graph as a collection of tuples rather than directly modeling the underlying semantic structure. 
Instead of considering graph generation as tuple detection, we build upon an exciting avenue of research in \textit{graph generative models}~\cite{liao2019efficient,belli2019image,ingraham2019generative,he2022td} to directly sample graph structures conditioned on images. By modeling the graph generation process as sequential decoding of adjacency lists, we can effectively model the interaction between detected entities using a simple, directed graph. A transformer-based relation classification model then converts the simple graph into a labeled, weighted, directed graph to generate scene graphs in an iterative, two-stage approach to move beyond triplet-based detection. 
\section{Proposed Approach}
\label{sec:ProposedApproach}
\begin{figure*}[th]
    \centering
    \includegraphics[width=0.99\textwidth]{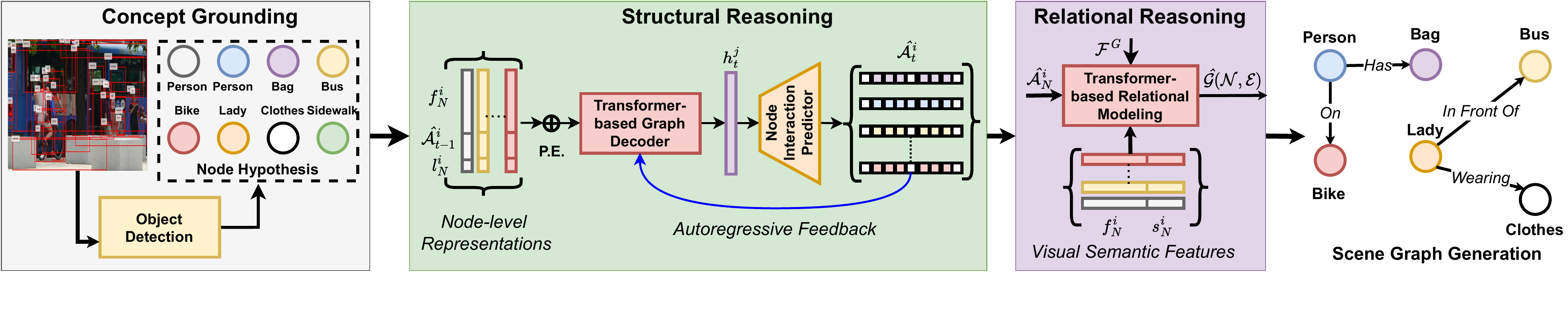}
    \caption{The \textbf{overall architecture} of the proposed IS-GGT is illustrated. 
    We first ground the concepts in the image data (Section~\ref{sec:grounding}) and use a generative transformer decoder network to sample an entity interaction graph (Section~\ref{sec:ggt}) before relation or predicate classification (Section~\ref{sec:rel_pred}) using a transformer-based contextualization mechanism for efficient scene graph generation.}
    \label{fig:arch}
\end{figure*}

\textbf{Overview.} We take a two-stage, generative approach to the problem of scene graph generation. The overall approach, called IS-GGT, is shown in Figure ~\ref{fig:arch}. There are three major components to the approach: (i) concept grounding, (ii) structural reasoning, and (iii) relational reasoning. Based on the idea of generative graph models, we use scene-level localization and entity concept hypothesis (Section~\ref{sec:grounding}) to first sample the underlying semantic structure of the scene using a generative transformer decoder network (Section~\ref{sec:ggt}). Once the semantic structure is sampled, the semantic relations (predicates), i.e., the edges, are labeled to characterize the scene graph (Section~\ref{sec:rel_pred}). 

\textbf{Problem Statement.} Scene graph generation (SGG) aims to generate a graph structure $\mathcal{G}=\{\mathcal{V}, \mathcal{E}\}$ from a given input image $I$, where $\mathcal{V}=\{v_1, v_2, \ldots v_n\}$ is the graph's nodes representing localized entities (objects) in the image and $\mathcal{E}=\{e_1, e_2, \ldots e_k\}$ represent the edges that describe the relationship connecting two nodes $n_i$ and $n_j$. Each node $v_i \in \mathcal{V}$ has two attributes, a label $l_i \in \mathcal{C_N}$ and a bounding box $bb_i$, where $\mathcal{C_N}$ is the space of all possible concepts in an environment. Each edge $e_i \in \mathcal{E}$ is characterized by a label $r_i \in \mathcal{R_K}$ and an optional assertion score $p(r_i)$, where $\mathcal{R}_K$ is the set of all possible relationships that can be present between the entities $\mathcal{C_N}$. 
Typical approaches to this problem have focused on extracting plausible triplets from an exhaustive search space consisting of all possible edges in a fully connected graph. Each node is connected to every other node. A relational prediction model is then trained to distinguish between the plausible relationship between the nodes, including \textit{null} relationship (indicated by a \textit{background} class). On the other hand, we aim to first sample the underlying semantic structure based on the node (entity) hypothesis to help model the global context before relationship classification. This allows us to reduce the computational overload for relationship prediction while restricting the relational reasoning to interactions that are considered to be plausible. 
We present the proposed framework below. 

\subsection{Concept Grounding: Entity  Hypotheses}\label{sec:grounding}
The scene graph generation process begins with entity hypotheses generation, which involves the localization and recognition of concepts in a given image $I$. Following prior work~\cite{zellers2018neural,Tang_2019_CVPR,xu2017scene}, we use a standard ResNet-based~\cite{he2016deep}, FasterRCNN~\cite{ren2015faster} model as the localization module. The object detector returns a set of $n$ detected entities $v_1, v_2, \ldots v_n$, characterized by their location using bounding boxes ($bb_1, bb_2, \ldots bb_n \in \mathcal{B}$) and corresponding labels ($l_1, l_2 \ldots l_n \mid l_i \in C_N$). These entities ($\mathcal{V}$) serve as our node hypothesis space, over which the scene graph generation is conditioned. Each entity is described by a feature representation ($f^i_N$) from the underlying ResNet encoder, through ROIAlign~\cite{he2017mask} using the predicted bounding boxes (ROIs) and the labels are generated through the classification layer from the object detector. Compared to prior work~\cite{zellers2018neural,Tang_2019_CVPR}, we do not have separate visual encoders for capturing the relationships among concepts at this stage. 
We allow the entities to be detected and represented independently, which enables us to decouple the ideas of graph prediction and predicate classification. 
%

\begin{algorithm}[t]
\caption{Scene semantic graph structure sampling using a generative transformer decoder.}
\label{alg:GGT}
\begin{algorithmic}[1]

\Require $\mathcal{V} = v_1, v_2, \ldots v_n \mid v_i = \{l_i, f^i_N, bb_i\}$
\Ensure $\mathcal{G} = \{\mathcal{V}, \mathcal{E}\} = \hat{\mathcal{A}_N} = \{\hat{\mathcal{A}^i_N}\}$
\State $ {\mathcal{G}} \gets \emptyset$ \Comment{\textit{Initialize empty graph}}
\State $ \hat{\mathcal{A}_N} \gets \emptyset$ \Comment{\textit{Initialize empty adjacency matrix}}
\State $\mathcal{E} \gets \emptyset$ \Comment{\textit{Initialize empty edge list}}
 \For{each node $v_i$ in $\mathcal{V}$}
    \State $c_t \gets [f^{1:i}_N; \hat{\mathcal{A}^{1:i}_N}, l_{1:i}]$\Comment{\textit{Context vector for decoding.}}
    \State $c_t \gets c_t + PositionalEncoding(c)$ 
    \State $h^0_t \gets MLP(c_t)$ \Comment{\textit{Linear projection}}
    \State $\hat{h}^K_l \gets TransformerDecoder(h^0_t)$ 
    \State $h^i_t \gets MLP(\hat{h}^K_l)$ \Comment{\textit{Learned feature space}}
    \State $\hat{\mathcal{A}^i_N} \gets Sample(\sigma(MLP(h_t))$ 
    \Comment{\textit{$v_i$'s adjacency list}}
    \State $\hat{l}_i \gets Softmax(MLP(h^i_t))$ \Comment{$v_i$'s auxiliary label}
    \State $\hat{\mathcal{A}_N} \gets \hat{\mathcal{A}_N} \bigcup \{\hat{\mathcal{A}^i_N}\}$ \Comment{\textit{Populate adjacency matrix}}
    \State $\mathcal{E} \gets \mathcal{E} \bigcup EdgeList(\hat{\mathcal{A}^i_N})$ \Comment{\textit{Collect edge list}}
 \EndFor
 \State $\mathcal{G} \gets \{\mathcal{V}, \mathcal{E}\}$ \Comment{\textit{Construct final interaction graph}}
\end{algorithmic}
\end{algorithm}

\subsection{Iterative Interaction Graph Generation}\label{sec:ggt}
At the core of our approach is the idea of graph sampling, where we first model the interactions between the detected entities in a graph structure. This sampled graph is a \textit{simple}, \textit{directed} graph, where the edges are present only between nodes (i.e., the detected entities) that share a semantically meaningful relationship. Each edge $e_i$ is unlabeled and merely signifies the plausible existence of a semantic relationship between the connecting nodes $v_i$ and $v_j$. Inspired by the success of prior work~\cite{belli2019image}, we model this graph generation process as the autoregressive decoding of the adjacency list $\mathcal{A}^i_N$ for each node $v_i$, using a transformer network~\cite{vaswani2017attention}. 
A simplified pseudocode of the whole process is shown in Algorithm~\ref{alg:GGT}. 
Given an empty graph $\mathcal{G}=\emptyset$, the underlying structural graph is generated through a sequence of edge and node additions. 
Each step of the decoding process emits an output adjacency list conditioned upon the visual features $f^i_N$ of each detected node $v_i$, its hypothesized label $l_i$ and the previously decoded adjacency matrices up to the current step $t$ given by $\hat{\mathcal{A}_t}$. 
This iterative graph generation process results in an adjacency matrix $\hat{\mathcal{A}} = \{\mathcal{A}^1_N, \mathcal{A}^2_N \ldots \mathcal{A}^n_N, \forall v_i \in \mathcal{V}\}$. 
The final adjacency matrix is an $N\times N$ matrix that can be sampled by some threshold $\gamma$ to obtain a binary adjacency matrix. The values where $\hat{\mathcal{A}}^i_N(i,j)=1$'s indicate that an edge is present between nodes $v_i$ and $v_j$, which can then be added to the edge list $\mathcal{E}$. The edge list is then sorted by its \textit{energy}, given by $E(e_{ij})=\sigma(p_i + p_j)$, where $p_i$ and $p_j$ refer to the confidence scores from the detector that provides a measure of confidence about the existence of the concepts $v_i$ and $v_j$ in the image, respectively. The collection of nodes $\mathcal{V}$ and edge list $\mathcal{E}$ provide the underlying semantic structure.

Formally, we define this process as maximizing the probability of observing a scene graph $\mathcal{G}$  conditioned on the input image $I$, and is given by
\begin{equation}
P(G \mid I) = P(\hat{\mathcal{A}}_N | I) = \prod_{i=1}^{N}{p(\hat{\mathcal{A}}^i_N \mid \hat{\mathcal{A}^{1:i}}_N, f^{1:i}_N, l_{1:i})}
\label{eqn:GGT_Objective}
\end{equation}
where we decompose the probability of observing the graph $\mathcal{G}$ as the joint probability over the separate adjacency lists for each node $v_i$ given its visual features $f^i_N$ and label $l_i$, along with the other nodes that have previously been sampled. Note that the ordering of the nodes can vary greatly; thus, search space to learn the sequence of adjacency lists can grow exponentially with the number of nodes. To this end, we present a fixed ordering of the nodes to be added to the graph based on the confidence score from the object detector to provide a tractable solution. We use a transformer-based decoder model trained in an auto-regressive manner to learn probability measure. 

The decoder is trained using two loss functions - an adjacency loss $\mathcal{L}_{\mathcal{A}}$ and a semantic loss $\mathcal{L}_{\mathcal{S}}$. The former is a binary cross-entropy loss between the predicted and actual binary adjacency matrix, while the latter is a cross-entropy loss for node label prediction. Specifically, we define $\mathcal{L}_{\mathcal{A}} = \frac{1}{N^2}\sum_{i=1}^N\sum_{j=1}^N{-(a_{ij}}log(\hat{a_{ij}}) + (1 - a_{ij})log(1 - \hat{a_{ij}}))$ and $\mathcal{L}_{\mathcal{S}} = -\sum_j^{\mathcal{C}}{l_j log(p(\hat{l}_j))}$, where $l_j$ is the entity's label as predicted by the concept grounding module from Section~\ref{sec:grounding} and $\hat{l}_j$ is the softmax probability from the node prediction of the transformer decoder as defined in line 11 of Algorithm~\ref{alg:GGT}. Note that we use the semantic loss $\mathcal{L}_{\mathcal{S}}$ as a mechanism to inject the semantics of the grounded concepts into the decoding process and do not use these predictions (termed \textit{node sampling}) as node labels for the final graph. We observe that node sampling (see Section~\ref{sec:ablation}) reduces the performance slightly. We attribute it to the fact the object detector has access to the global, image-level context and hence has a better classification performance. 
The total loss is given by
\begin{equation}
\mathcal{L}_{G} = \lambda \mathcal{L}_{\mathcal{A}} + (1-\lambda) \mathcal{L}_{\mathcal{S}}
    \label{eqn:GGT_loss}
\end{equation}
where $\lambda$ is a trade-off between semantic and adjacency losses. In our experiments, we set $\lambda=0.75$ to place more emphasis on the adjacency loss. During training, we use teacher forcing in the transformer decoder and convert the adjacency matrix to binary for tractable optimization. 

\begin{table*}[ht]
\centering
\resizebox{0.99\textwidth}{!}{
\begin{tabular}{|c|c|c|c|c|c|c|c|c|c|}
\hline
 \multicolumn{1}{|c|}{} & \multirow{2}{*}{{\textbf{Approach}}} & \multicolumn{2}{c|}{\textbf{PredCls}} & 
    \multicolumn{2}{c|}{\textbf{SGCls}} & \multicolumn{2}{c|}{\textbf{SGDet}} & \multicolumn{1}{p{1.2cm}|}{\textbf{Average}} & \multicolumn{1}{p{1.2cm}|}{\textbf{Average}}\\
\cline{3-8}
 &  & \textbf{mR@50} & \textbf{mR@100} & \textbf{mR@50} & \textbf{mR@100} & \textbf{mR@50} & \textbf{mR@100} & \textbf{mR@100} &  \textbf{mR@50}\\
\hline
 \multirow{10}{*}{\rotatebox[origin = c]{90}{\textbf{Without Unbiasing}} }
 & FC-SSG~\cite{Liu_2021_CVPR} & 6.3 & 7.1 & 3.7 & 4.1 & 3.6 & 4.2 & 4.5 & 5.1 \\
 & IMP\cite{xu2017scene} & 9.8 & 10.5 & 5.8 & 6.0 & 3.8 & 4.8 & 7.1 & 6.5\\
 & MOTIFS \cite{zellers2018neural} & 14.0 & 15.3 & 7.7 & 8.2 & 5.7 & 6.6 & 10.0 & 9.1 \\
 & VCTree\cite{Tang_2019_CVPR} & 17.9 & 19.4 & 10.1 & 10.8 & 6.9 & 8.0 & 12.7 & 11.6 \\
 & KERN\cite{Chen_2019_CVPR} & - & 19.2 & - & 10 & - & 7.3 & 12.2 & - \\
&  R-CAGCN\cite{yang2021probabilistic} & - & 19.9 & - & 11.1 & - & 8.8 & 13.3 & - \\
&  Transformer\cite{guo2021general} & - & 17.5 & - & 10.2 & - & 8.8 & 12.2 & - \\
&  Relationformer\cite{shit2022relationformer} & - & - & - & - & \underline{9.3} & 10.7 & - & - \\
&  RelTR\cite{cong2022reltr} & 21.2 & - & 11.4 & - & 8.5 & - & - & 13.7 \\
\cmidrule{2-10}

&  \textbf{IS-GGT (Ours)} & \textbf{26.4} & \textbf{31.9} & \textbf{15.8} & \textbf{18.9} & \textbf{9.1} & \textbf{11.3} & \textbf{20.7} & \textbf{17.1} \\
\midrule
\midrule
\multirow{15}{*}{\rotatebox[origin = c]{90}{\textbf{With Unbiasing}} } 
 & RU-Net\cite{Lin_2022_CVPR} & - & 24.2 & - & 14.6 & - & 10.8 & 16.5 & -\\

 \cmidrule{2-10}

 & IMP+EBML\cite{suhail2021energy} & 11.8 & 12.8 & 6.8 & 7.2 & 4.2 & 5.4 & 8.46 & 7.6 \\
 & VCTree+EBML\cite{suhail2021energy} & 18.2 & 19.7 & 12.5 & 13.5 & 7.7 & 9.1 & 14.1 & 12.8\\
 & MOTIFS+EBML\cite{suhail2021energy} & 18.0 & 19.5 & 10.2 & 11 & 7.7 & 9.1 & 13.2 & 12.0\\

 \cmidrule{2-10}

 & MOTIFS+TDE\cite{tang2020unbiased} & {25.5} & 29.1 & 13.1 & 14.9 & 8.2 & 9.8 & 17.9 & 15.6\\
 & VCTree+TDE\cite{tang2020unbiased}& {25.4} & 28.7 & 12.2 & 14 & \underline{9.3} & {11.1} & 17.9 & 15.6\\

 \cmidrule{2-10}
 
 & MOTIFS+CogTree\cite{yu2020cogtree} & {26.4} & 29 & 14.9 & 16.1 & \underline{10.4} & \underline{11.8} & 19.0 & \underline{17.2}\\
 & VCTree+CogTree\cite{yu2020cogtree} & \underline{27.6} & 29.7 & \underline{18.8} & \underline{19.9} & \underline{10.4} & \underline{12.1} & {20.6} & \underline{18.9}\\

 \cmidrule{2-10}

 & IMP+PPDL\cite{Li_2022_CVPR} & {24.8} & 25.3 & 14.2 & 15.9 & \underline{9.8} & 10.4 & 17.2 & 16.2\\
 & MOTIFS+PPDL\cite{Li_2022_CVPR} & \underline{32.2} & \underline{33.3} & 17.5 & 18.2 & \underline{11.4} & \underline{13.5} & \underline{21.7} & \underline{20.4}\\
 & VCTree+PPDL\cite{Li_2022_CVPR} & \underline{33.3} & \underline{33.8} & \underline{21.8} & \underline{22.4} & \underline{11.3} & \underline{14.4} & \underline{23.5} & \underline{22.1}\\

\cmidrule{2-10}

 & BGNN\cite{Li_2021_CVPR} & \underline{30.4} & \underline{32.9} & 14.3 & 16.5 & \underline{10.7} & \underline{12.6} & {20.7} & \underline{18.5}\\
 
 \cmidrule{2-10}

 & PCPL\cite{10.1145/3394171.3413722} & \underline{35.2} & \underline{37.8} & \underline{18.6} & \underline{19.6} & \underline{9.5} & \underline{11.7} & \underline{23.0} & \underline{21.1}\\
\hline
\end{tabular}
}
\caption{\textbf{Comparison with the state-of-the-art} scene graph generation approaches, with and without unbiasing. We consistently outperform all models that do not use unbiasing and some early unbiasing models across all three tasks while offering competitive performance to current state-of-the-art unbiasing models. Approaches outperforming the proposed IS-GGT are underlined.}
\label{tab:sota}
\end{table*}

\subsection{Edge Labeling: Relation Prediction}\label{sec:rel_pred}
The final step in the proposed approach is predicate (or entity relation) prediction, which involves the labeling of the edges $\mathcal{E}$ in the interaction graph $\mathcal{G}$ generated from Section~\ref{sec:ggt}. To further refine the interaction graph, we assign an ``edge prior'' to each sampled edge $e_{ij} \in \mathcal{E}$ between two nodes $n_i$ and $n_j$. This prior is a function of the confidence scores ($c_i$ and $c_j$, respectively) obtained from the concept grounding module (Section~\ref{sec:grounding}) and is given by $E(e_{ij}) = \sigma(c_i \times c_j)$. Finally, we sort the edges based on their edge prior and take the top $K$ edges as the final edge list to represent the scene graph $\mathcal{G}_s$. In our experiments, we set $K{=}250$ to provide a tradeoff between inference time and expressiveness, although we find that lower values of $K$ do not reduce the performance (see Section~\ref{sec:gg_eval}). Given the final edge list $\mathcal{E}$, we then predict the relationship by maximizing the probability $P(r_{k} \mid f^i_N, f^j_N, S^i_N, S^j_N, bb_i, bb_j, \mathcal{F}^G)$, where $\mathcal{F}^G$ is the global image context captured by a contextualization mechanism, and $r_{k}$ is the relationship of the $k^{th}$ edge between nodes $n_i$ and $n_j$ described by their visual features $f^i_N$ and $f^j_N$, and semantic features $S^i_N$ and $S^j_N$, respectively. 
We obtain the contextualized global features $\mathcal{F}^G$ using DETR~\cite{carion2020end}. 
The semantic features are obtained through an embedding layer initialized by pre-trained word embeddings of the concept labels $\mathcal{C}$ such as GloVe~\cite{pennington2014glove} or ConceptNet Numberbatch~\cite{speer2017conceptnet}. We use an encoder-decoder transformer~\cite{vaswani2017attention} to model this probability. 
Specifically, we use a linear projection to map the entity features (visual features $F^i_N$ and localization features $bb_i$) of each node in the edge $e_k = e_{ij} \in \mathcal{E}$ into a shared visual embedding space by $\hat{h
}^k_v=RELU(W_c [f^i_N; bb_i; f^j_N; bb_j])$. A visual-semantic entity embedding is obtained by a linear projection and is given by $\hat{h}^k_{sv} = RELU(W_{sv} [\hat{h}^k; S^i_N, S^j_N])$. An encoder-decoder transformer then takes these visual-semantic features to predict the relationship through a series of attention-based operations given by
\begin{align}
    h^k_{sv} &= {Att}^E_{enc}(Q = K = V = \hat{h}^k_{sv})\\
    \hat{h}^k &= {Att}^D_{dec}(Q = \hat{h}^k_{sv}, K = V = \mathcal{F}^G)
    \label{eqn:transformer}
\end{align}
where ${Att}^E_{enc}(\dots)$ is a transformer encoder consisting of $E$  multi-headed attention layer ($MHA(Q,K,V) {=} W_{a}[h_1; h_2; \ldots h_K]$), as proposed in Vaswani \textit{et al.}~\cite{vaswani2017attention}, where $h_i{=}Attn(Q{=}W_QX, K{=}W_KX, V{=}W_VX)$. The multi-headed attention mechanism applies a scaled dot product attention operation given by  $Attn(Q,K,V){=}Softmax(\frac{QK^T}{\sqrt{D_K}}V)$. The resulting vector $h^k_{sv}$ is then passed through a D-layer transformer decoder that obtains a contextualized representation $\hat{h}^k$ for each edge $e_k$ with respect to the global context $\mathcal{F}^G$. The relationship (or predicate) for each edge is obtained by applying a linear layer on $\hat{h}_k$ followed by softmax to obtain the probability of a relationship $p(\hat{r}_k)$. We train this network using a weighted cross-entropy loss given by 
\begin{equation}
\mathcal{L}_{R} = -w_r\sum_{l=1}^{\mathcal{C_N}}{r_klog(\hat{r}_k)}
    \label{eqn:cross_entropy}
\end{equation}
where $r_k$ is the target relationship class, $\hat{r}_k$ is the probability of the predicted relationship class and $w_r$ is the weight given to correct relationship class. In our experiments, we set the weights as the inverse of the normalized frequency of occurrence of each relationship $r_k\in \mathcal{C_N}$. The weighted cross-entropy allows us to address the long-tail distribution of the predicate relationships in the scene graph classification task in a simple yet efficient manner.

\textbf{Implementation Details.} 
In our experiments, we use a Faster RCNN model with ResNet-101~\cite{he2016deep} as its backbone, trained on Visual Genome, and freeze the detector layers. The features extracted from the object detector were $2048$ dimensions and were filtered to obtain bounding boxes specific to the target vocabulary. The iterative graph decoder from Section~\ref{sec:ggt} has a hidden size of dimension $256$ and 6 layers with a sinusoidal positional encoding and is trained for 50 epochs with a learning rate of $0.001$. The predicate classifier (Section~\ref{sec:rel_pred}) is set to have $256$ in its hidden state for both networks, and GloVe embeddings~\cite{pennington2014glove} with 300-d vectors are used to derive the semantic features $S^i_N$. The predicate classifier is trained for $20$ epochs with a learning rate of $1\times 10^{-4}$. The training took around 3 hours for both networks on a GPU server with a 64-core AMD Threadripper processer and 2 NVIDIA Titan RTX GPUs.

\begin{table}[t]
    \centering
    \resizebox{0.475\textwidth}{!}{
    \begin{tabular}{|c|c|c|c|c|}
    \toprule
         \multirow{2}{*}{\textbf{Approach}} & \textbf{PredCls} & \textbf{SGCls} & \textbf{SGDet} & \textbf{Mean}\\
          & \textbf{zR@\{20/50\}} & \textbf{zR@\{20/50\}} & \textbf{zR@\{20/50\}} & \textbf{\textbf{zR@\{20/50\}}}\\
    \midrule
    VCTree\cite{Tang_2019_CVPR} & 1.4 / 4.0 & 0.4 / 1.2 & 0.2 / 0.5 & 0.7 / 1.9 \\
    MOTIFS\cite{zellers2018neural} & 1.3 / 3.6 & 0.4 / 0.8 & 0.0 / 0.4 & 0.6 / 1.7 \\
    FC-SGG~\cite{Liu_2021_CVPR} & -/\underline{7.9} & -/1.7 & -/\underline{0.9} & -/\underline{3.5}\\
    VCTree + EBML\cite{suhail2021energy} & \underline{2.3} / {5.4} & \underline{0.9} / \underline{1.9} & \underline{0.2} / {0.5} & \underline{1.1} / {2.6} \\
    MOTIFS + EBML\cite{suhail2021energy} & 2.1 / 4.9 & 0.5 / 1.3 & 0.1 / 0.2 &  0.9 / 2.1 \\ 
    IS-GGT (Ours) & \textbf{5.0} / \textbf{8.3} & \textbf{1.4} / \textbf{2.6} & \textbf{1.0} / \textbf{1.3} &  \textbf{2.5} / \textbf{4.1}\\
    \bottomrule
    \end{tabular}
    }
    \caption{\textbf{Zero-shot evaluation} on Visual Genome. We report the recall@20 and recall@50 for fair comparison.}
    \label{tab:zeroshot}
\end{table}

\section{Experimental Evaluation}
\label{sec:eval}
\textbf{Data.} We evaluate our approach on Visual Genome~\cite{krishna2017visual}. Following prior works~\cite{zellers2018neural,xu2017scene,Tang_2019_CVPR,Chen_2019_CVPR}, we use the standard scene graph evaluation subset containing 108k images with 150 object (entity) classes sharing 50 types of relationships (predicates). We use the $70\%$ of the data for training, whose subset of 5,000 images is used for validation, and the remaining 30\% is used for evaluation. 
%
We evaluate our approach on three standard scene graph generation tasks - predicate classification (\textbf{PredCls}), scene graph classification (\textbf{SGCls}), and scene graph generation (\textbf{SGDet}). The goal of PredCls is to generate the scene graph, given ground truth entities and localization, while in SGCls, the goal is to generate the scene graph, given only entity localization. In SGDet, only the input image is provided, and the goal is to generate the scene graph along with the entity localization.

\textbf{Metrics and Baselines.} Following prior work~\cite{Chen_2019_CVPR,Tang_2019_CVPR,cong2022reltr,yang2021probabilistic}, we report the mean recall (\textbf{mR@K}) metric, since the recall has shown to be biased towards predicate classes with larger amounts of training data~\cite{Chen_2019_CVPR,Tang_2019_CVPR}. We report across different values of $K \in. \{50,100\}$ We also present the average mR@K across all tasks to summarize the performance of the scene graph generation models across the three tasks with varying difficulty. We also report the zero-shot recall (\textbf{zsR@K}, $K\in\{20,50\}$) to evaluate the generalization capabilities of the SGG models. Finally, we compare against two broad categories of scene graph generation models - those with unbiasing and those without unbiasing. Unbiasing refers to the use of additional training mechanisms, such as leveraging prior knowledge to tackle the long-tail distribution in predicate classification. All numbers are reported under the \textit{with graph constraint} setting. 

\begin{figure}[t]
    \centering
    \includegraphics[width=0.95\columnwidth]{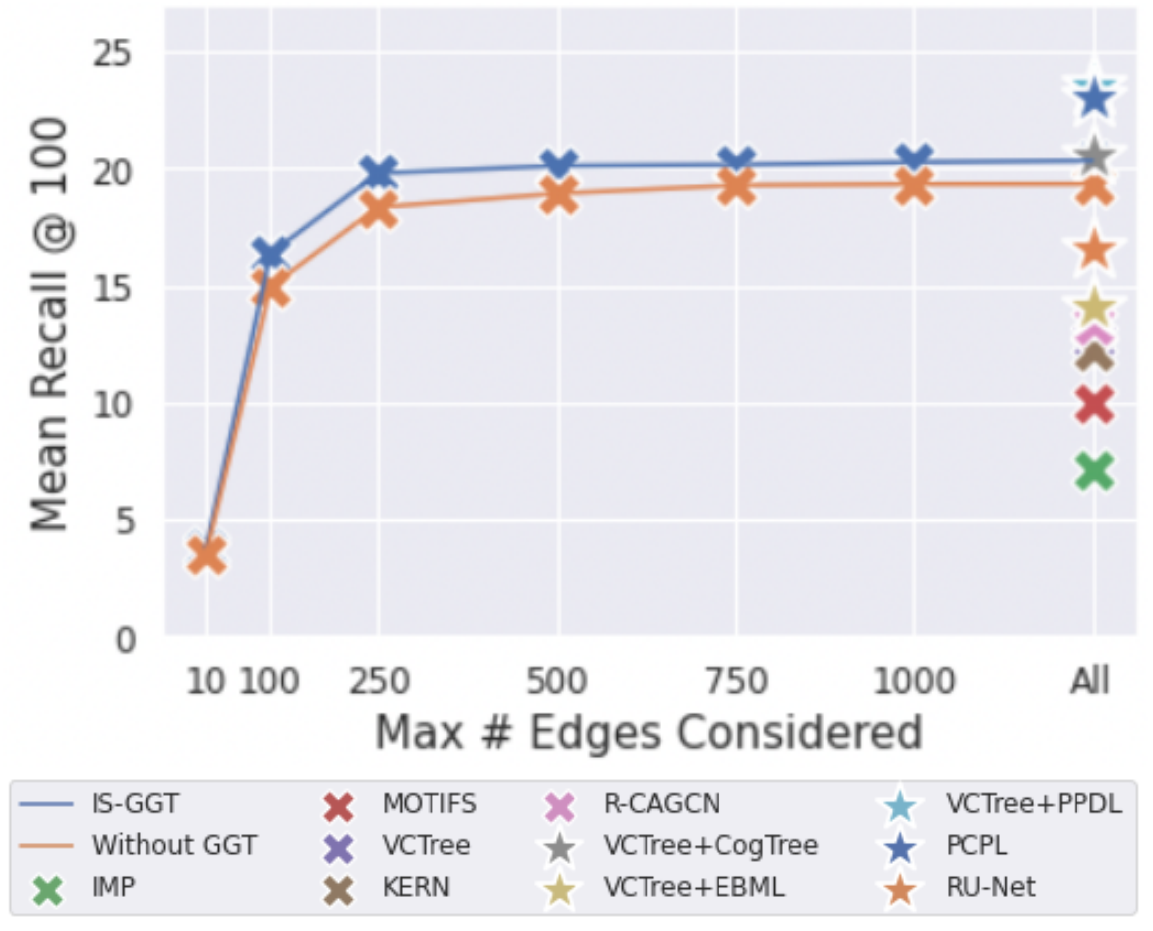}
    \caption{\textbf{Impact of graph sampling.} 
    We greatly reduce the number of pairwise comparisons made for scene graph generation. 
    Using only 200 edges ($\sim 20\%$ of all edges), we outperform most state-of-the-art approaches on the mean mR@100 across all tasks.}
    \label{fig:GGT_plot}
\end{figure}

\subsection{Comparison with State-Of-The-Art}\label{sec:SOTA}
We evaluate our approach on the test split of Visual Genome with the mean recall under graph constraints metric (mR@50 and mR@100) and compare with several state-of-the-art scene graph generation approaches, both with and without unbiasing. The results are summarized in Table~\ref{tab:sota}. Without bells and whistles, we significantly outperform approaches that do not use unbiasing across all three tasks. Interestingly, we outperform the closely related, transformer-based ReITR~\cite{cong2022reltr} model by $2.7$ points in the average mR@50 metric. 
In comparison with models with unbiasing, we see that we perform competitively to current state-of-the-art models such as PPDL~\cite{Li_2022_CVPR}, CogTree~\cite{yu2020cogtree}, and BGNN~\cite{Li_2021_CVPR}, while outperforming some of the earlier approaches to unbiasing such as EBML~\cite{suhail2021energy} and TDE~\cite{tang2020unbiased} across all tasks. 
Of particular interest is the comparison with RU-Net~\cite{Lin_2022_CVPR}, a scene graph generation model that jointly models unbiasing and generation in a unified framework, as opposed to other approaches, which primarily focus on improving the predicate classification performance of underlying SGG models. We consistently outperform RU-Net across all three tasks, with an average mR@100 improvement of $3.6$ absolute points. It is also remarkable to note the performance difference (in mR@100) between the state-of-the-art unbiasing model (PPDL) and our IS-GGT on PredCls is less than $3\%$, considering that they are optimized specifically for this task, indicating that the graph sampling approach consistently places the edges in the ground truth scene graph in the top 100 edges. 

\begin{table}[t]
    \centering
    \resizebox{0.475\textwidth}{!}{
    \begin{tabular}{|c|c|c|c|c|}
    \toprule
         \textbf{Max Edges} & \textbf{PredCls} & \textbf{SGCls} & \textbf{SGDet} & \textbf{Graph Acc.}\\
         \textbf{Considered} & \textbf{mR@100} & \textbf{mR@100} & \textbf{mR@100} & \textbf{unconst. (\textit{const.})}\\
    \midrule
    10 & 4.6 & 3.3 & 3.5 &  11.6 (\textit{9.1})\\
    100 & 24.3 & 14.0 & 10.8 & 35.1 (\textit{25.3})\\
    250 & 30.1 & 17.5 & 11.8 & 44.2 (\textit{30.7})\\
    500 & 30.8 & 17.6 & 11.9 & 49.5 (\textit{33.3})\\
    750 & 31.0 & 17.6 & 11.9 & 51.4 (\textit{34.4})\\
    All & 31.4 & 17.6 & 12.0 & 52.7 (\textit{34.8})\\
    \bottomrule
    \end{tabular}
    }
    \caption{The \textbf{quality of the sampled edges} is quantified using its impact on the three scene graph generation tasks. mR@100 and average graph accuracy are reported.}
    \label{tab:ggt_impact}
\end{table}

\textbf{Zero-Shot Evaluation.} We also evaluated the generalization capabilities of our approach by considering the zero-shot evaluation setting. Here, the recall (with graph constraint) was computed only on edges (i.e., subject-predicate-object pairs) that were not part of the training set and summarize the results in Table~\ref{tab:zeroshot}. It can be seen that we outperform approaches with and without unbiasing. Specifically, we obtain and average zero-shot recall of 2.2 (at $K{=}20$) and 4.0  (at $K{=}50$), which is more than $2\times$ the performance of comparable models without unbiasing such as VCTree and MOTIFS while also outperforming the comparable FC-SGG~\cite{Liu_2021_CVPR} across all three tasks. It is interesting to note that we also outperform EBML~\cite{suhail2021energy}, which proposes to mitigate the long-tail distribution using an energy-based loss function. Interestingly, our approach, IS-GGT obtains $21.4$ zR@100, \textit{without graph constraint}, which outperforms FC-SGG~\cite{Liu_2021_CVPR} ($19.6$), VCTree+TDE~\cite{tang2020unbiased} ($17.6$), and MOTIFS+TDE~\cite{tang2020unbiased} ($18.2$) which are state-of-the-art unbiasing models in the zero-shot regimen. 

\subsection{Importance of Graph Sampling.}\label{sec:gg_eval}
At the core of our approach is the notion of graph sampling, as outlined in Section~\ref{sec:ggt}. Hence, we examine its impact on the performance of the proposed IS-GGT in more detail. First, we assess the effect of considering the top K edges based on the edge prior (Section~\ref{sec:rel_pred}), which directly impacts the number of edges considered in the final graph for predicate classification. 
We vary the maximum number of edges considered per predicted scene graph from 10 to 1000 and consider all pairwise comparisons for each detected entity. We assess its impact on the average mean recall (mR@100) across all three tasks (PredCls, SGCls, and SGDet) and summarize the result in Figure~\ref{fig:GGT_plot}. As can be seen, we outperform all SGG models that do not use unbiasing while considering only the top $100$ edges, which represents $\sim 10\%$ of all possible pairwise combinations while at $K{=}200$ edges outperform most models \textit{with} unbiasing. Only PCPL~\cite{10.1145/3394171.3413722} and PPDL~\cite{Li_2022_CVPR} outperform IS-GGT, although they consider all (($>1000$) combinations. 

\begin{table}[t]
    \centering
    \resizebox{0.475\textwidth}{!}{
    \begin{tabular}{|c|c|c|c|c|c|c|}
        \toprule
        \textbf{G.C.} & \textbf{V.F.} & \textbf{S.F.} & \textbf{G.S.} & \textbf{PredCls} & \textbf{SGCls} & \textbf{SGDet}\\
        \toprule
        \ding{51} & \ding{51} & \ding{55} & \ding{51} & 28.3 & 16.5 & 10.3 \\ 
        \ding{51} & \ding{51} & {C.N.B.} & \ding{51} & 28.5 & 16.8 & 11.6 \\ 
        \ding{51} & \ding{51} & {GloVe} & \ding{51} & \textbf{30.1} & \textbf{17.4} & \textbf{11.9} \\ 
        \midrule
        \ding{51} & \ding{55} & {GloVe} & \ding{51} & 29.2 & 15.2 & 10.0 \\ 
        \ding{55} & \ding{51} & {C.N.B} & \ding{51} & 28.5 & 16.9 & 11.0 \\ 
        \ding{55} & \ding{51} & {GloVe} & \ding{51} & 29.3 & 16.9 & 10.5 \\ 
        \midrule
        \ding{55} & \ding{51} & {GloVe} & \ding{55} & 27.9 & 16.1 & 11.0 \\ 
        \ding{51} & \ding{51} & {GloVe} & \ding{55} & 28.5 & 16.8 & 11.2 \\ 
        \ding{51} & \ding{51} & {GloVe} & W/o E.P. & N/A & 17.2 & 9.3 \\ 
        \ding{51} & \ding{51} & {GloVe} & With N.S. & 28.5 & 17.2 & 8.9\\
        \bottomrule
    \end{tabular}
    }
    \caption{\textbf{Ablation studies} are presented to quantify each component's impact on mR@100. G.C.: global context, V.F.: visual features, S.F: semantic features, G.S.: graph sampling, C.N.B: ConceptNet Numberbatch, E.P. edge prior, and N.S: node sampling.
    } 
    \label{tab:ablation}
\end{table}

\begin{figure*}
    \centering
    \begin{tabular}{cc}
    \toprule
    \multicolumn{2}{c}{\textbf{Scene Graph Detection}}\\
    \toprule
    \includegraphics[width=0.5\textwidth]{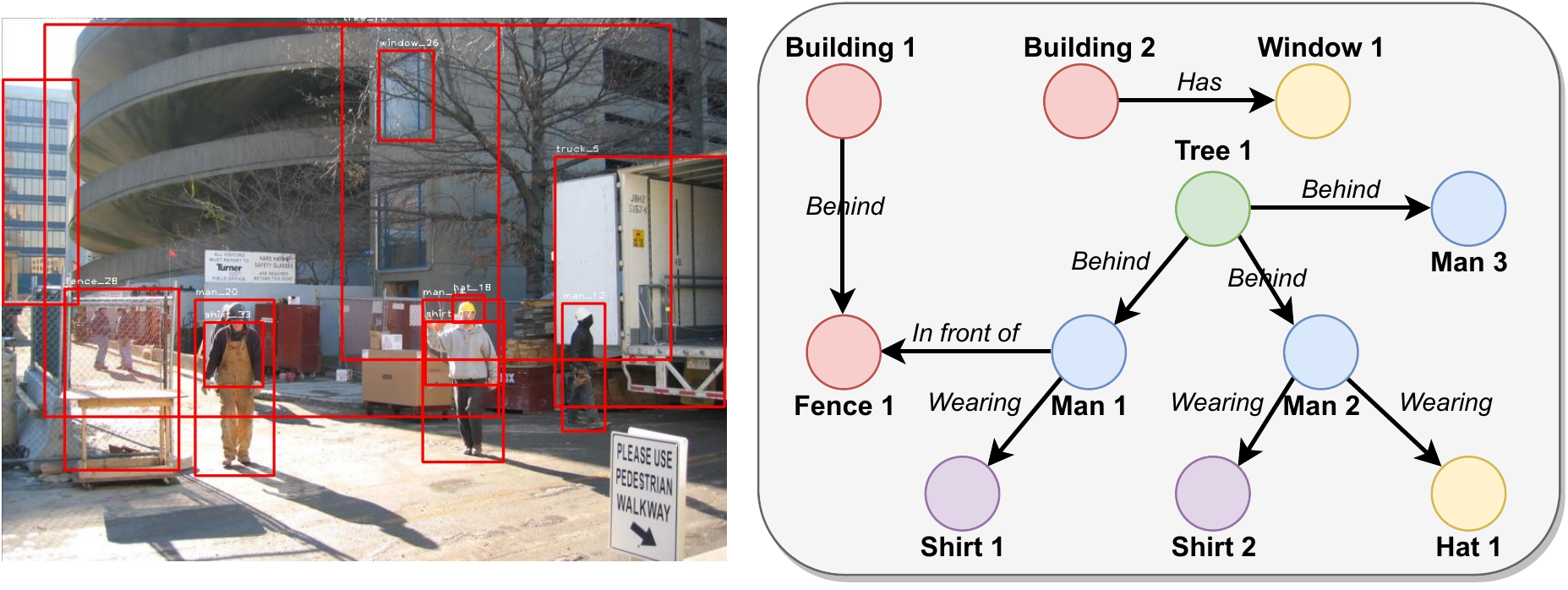}     &  
    \includegraphics[width=0.43\textwidth]{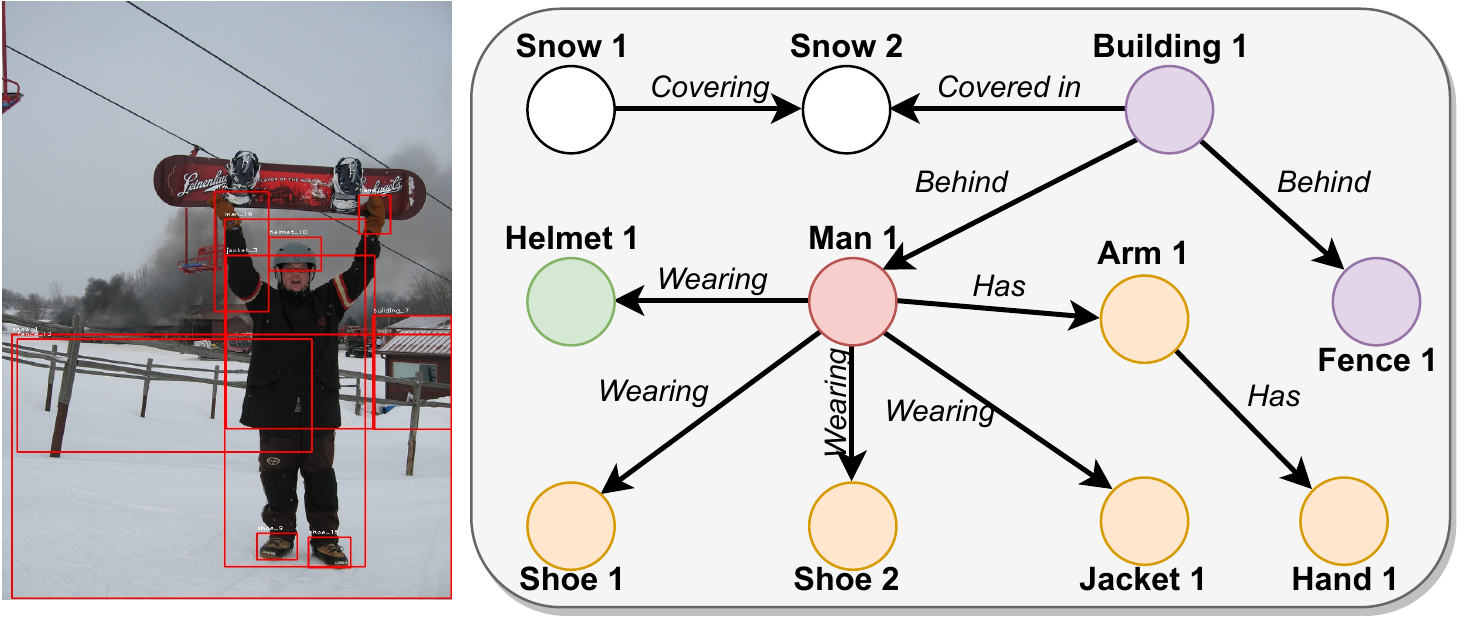} \\
    (a) & (b) \\
    \midrule
    \multicolumn{1}{c}{\textbf{Predicate Classification with Zero-Shot Edges}} & 
    \multicolumn{1}{c}{\textbf{Predicate Classification}}\\
    \midrule
    \includegraphics[width=0.52\textwidth]{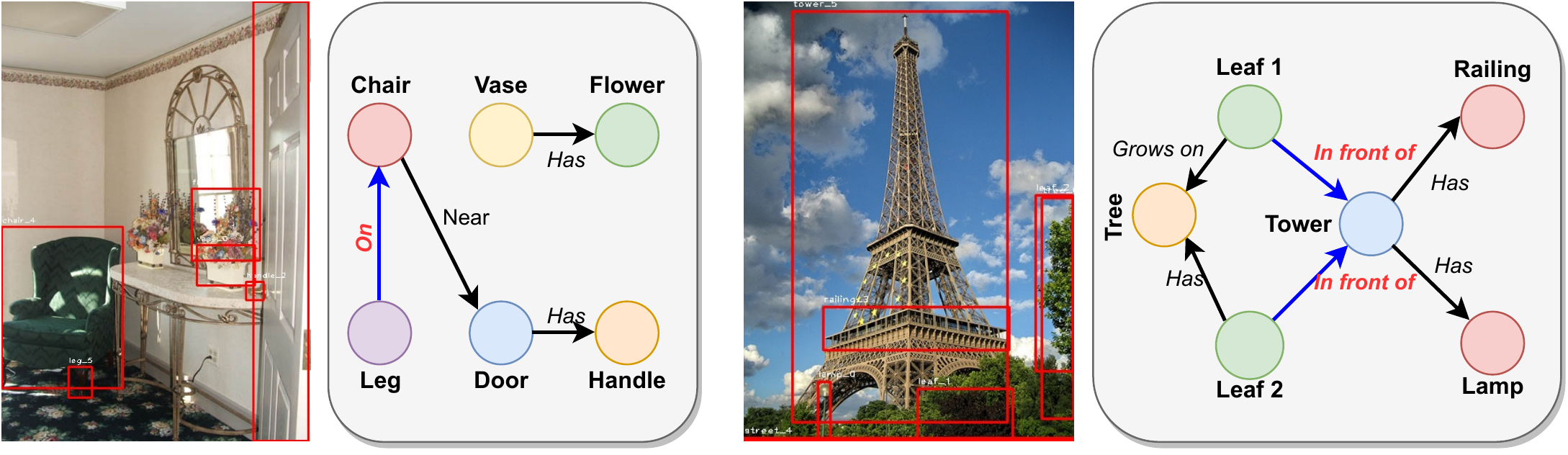} & 
    \includegraphics[width=0.43\textwidth]{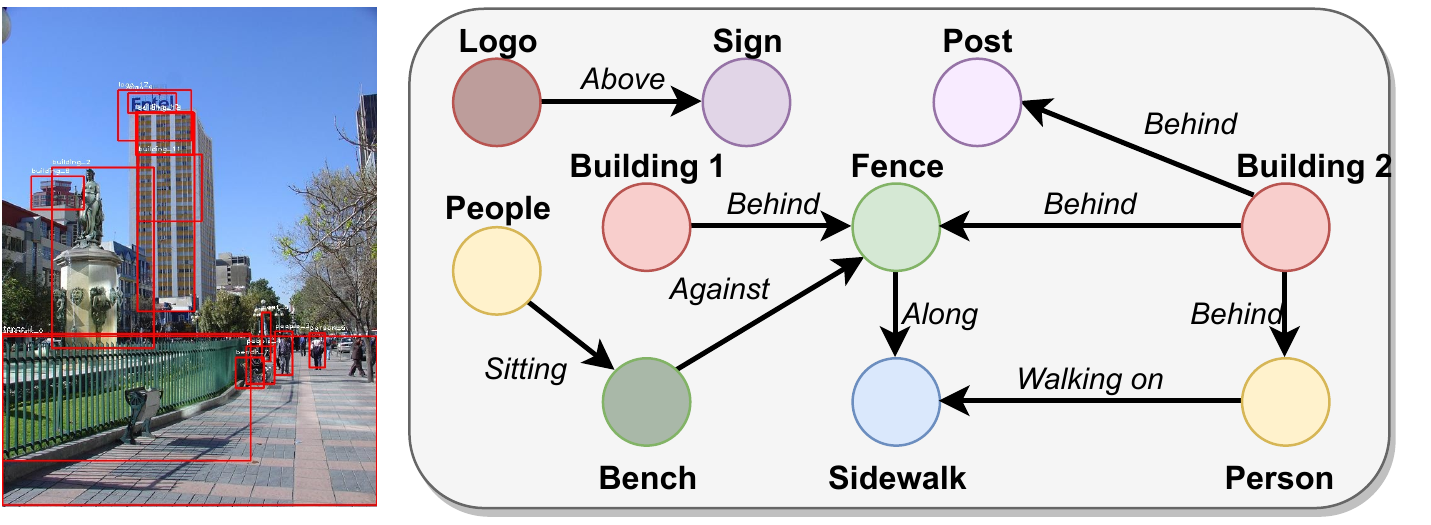} \\
    (c) & (d) \\
    \bottomrule
    \end{tabular}
    \caption{We present \textbf{qualitative visualizations} of the scene graphs generated by IS-GGT under (a) scene graph detection setting, (b) predicate classification on images with zero-shot predicates (indicated by blue edges), and (c) predicate classification with complex structures.}
    \label{fig:qual}
\end{figure*}

In addition to the impact on the average mR@100, we also assess the \textit{quality} of the underlying graph sampled with the generative graph transformer decoder. We propose two new metrics, unconstrained and constrained graph accuracy, which measure the quality of the sampled edges. In the former, we measure the accuracy of the underlying structure by when both the nodes and edges are unlabeled and binary. In the latter, we only consider the edges to be unlabeled. Note that, in both metrics, for a node to be ``correct'', its bounding box must have at least $50\%$ overlap with a corresponding ground truth node. We summarize the results in Table~\ref{tab:ggt_impact}. It can be seen that the graph accuracy increases with the number of considered edges while plateauing out at around $500$ edges. Interestingly, the constrained accuracy, IS-GGT's theoretical upper bound, is $30.7\%$ with only $250$ sampled edges. 
This is a remarkable metric considering that, on average. the number of total possible edges per image can be more than $1000$, and more than $30\%$ of the ground truth edges are part of the top $250$ edges. 
These results indicate that the graph sampling does an effective job in capturing the underlying entity interaction structure. 

\subsection{Ablation Studies}\label{sec:ablation}
To assess the impact of each component in the proposed IS-GGT framework, we systematically evaluate the framework's performance by exploring alternatives, including the exclusion of each element. Specifically, we assess the impact of three broad categories - (i) use of semantics, (ii) choice of visual features, and (iii) use of graph sampling. 
We see that the lack of semantic features has a more significant impact, resulting in a reduction of an average of $1.47\%$ in absolute mR@100 across tasks. In contrast, the choice of semantic features (ConceptNet Numberbatch~\cite{speer2017conceptnet} vs. GloVe~\cite{pennington2014glove}) has limited impact. 
We attribute the success of GloVe to its pre-training objective, which ensures that the dot product between GloVe embeddings is proportional to their co-occurrence frequency. This property helps model the potential semantic relationships between nodes using the attention mechanism in relationship prediction model (Section~\ref{sec:rel_pred}). 
Interestingly, we see that adding global context as part of the predicate prediction features (Section~\ref{sec:rel_pred}) significantly improves the performance ($\sim 1.1\%$ average mR@100), whereas removing visual context altogether also results in a reduction of $\sim 1.7\%$ average mR@100. 
Removing the GGT and removing the edge prior also hurt the performance. However, the recall metric does not accurately capture the reduction in the false alarms produced due to the lack of edge sampling with a generative model. Finally, we see that using node sampling ($\hat{l}_i$ from Section~\ref{sec:ggt}) affects SGCls and SGDet significantly. 
We attribute it to the fact that concept grounding is an essential step in modeling the visual-semantic relationships among entities.

\textbf{Qualitative Evaluation.} We present some qualitative illustrations of some of the scene graphs generated by the proposed approach in Figure~\ref{fig:qual}. In the top row, we present the generated scene graphs under the ``detection'' setting, where the goal is to both detect entities and characterize the relationships between them. It can be seen that, although there are a large number of detected entities ($\sim 28$ per image), the graph sampling approach allows us to reject clutter to arrive at a compact representation that captures the underlying semantic structure. Figure~\ref{fig:qual}(c) shows the generalization capabilities of the proposed approach for predicate classification when previously unseen (``zero-shot'') triplets are observed. Finally, we show in Figure~\ref{fig:qual}(d) that the graph sampling also works under cluttered scenarios when localized, ground-truth entities are provided, and there is a need to reject nodes that do not add to the scene's semantic structure. We can sample sparse graph structures that express complex semantics without losing expressiveness.

\section{Conclusion}
\label{sec:conclusion}
In this work, we presented IS-GGT, one of the first works to address the problem of generative graph sampling for scene graph generation. Using a two-stage approach, we first sample the underlying semantic structure of the scene before predicate (relationship) characterization. This decoupled prediction allows us to reason over the constrained (optimal) global semantic structure while reducing the number of pairwise comparisons for predicate classification. Extensive experiments on visual genome indicate that the proposed approach outperforms scene graph generation models without unbiasing while offering competitive performance to those with unbiasing while considering only $\sim 20\%$ of the total possible edges. We aim to extend this approach for general graph generation problems such as semantic graphs~\cite{aakur2019going} and temporal graph prediction~\cite{bal2022bayesian,ji2020action}, where capturing the underlying entity interactions can help constrain the search space for complex reasoning. 

\textbf{Acknowledgements.} This research was supported in part by the US National Science Foundation grants IIS 2143150, and IIS 1955230.

{\small
\bibliographystyle{ieee_fullname}
\bibliography{egbib}
}

\end{document}